\documentclass{ws-procs9x6}

\begin{document}

\title{On the possible Computational Power\\ of the Human Mind}

\author{HECTOR ZENIL\footnote{zenil@ciencias.unam.mx}  ~ \& FRANCISCO HERNANDEZ-QUIROZ\footnote{fhq@hp.fciencias.unam.mx
}}
\address{Math Department, Faculty of Science,\\ National University of Mexico (UNAM)
\linebreak \\ Published in WORLDVIEWS, SCIENCE AND US, edited by Carlos Gershenson, Diederik Aerts and Bruce Edmonds, World Scientific, 2007.}
\maketitle

\abstracts{The aim of this paper is to address the question: Can an artificial neural network (ANN) model be used as a possible characterization of the power of the human mind? We will discuss what might be the relationship between such a model and its natural counterpart. A possible characterization of the different power capabilities of the mind is suggested in terms of the information contained (in its computational complexity) or achievable by it. Such characterization takes advantage of recent results based on natural neural networks (NNN) and the computational power of arbitrary artificial neural networks (ANN). The possible acceptance of neural networks as the model of the human mind's operation makes the aforementioned quite relevant.}

\section{Introduction}

Much interest has been focused on the comparison between the brain and computers. A variety of obvious analogies exist. Based on several thoughts, some authors from very diverse research areas (Philosophy, Physics, Computer Science, etc.) claim that the human mind could be more powerful than Turing Machines\cite{penrose3,searle,copeland2,Sieg2}. Nevertheless, they do not agree on what these ``super-Turing'' capabilities mean. Consequently, there is no universally accepted characterization of this extra power and how it could be related to the human mind, even though there is strong defense of these authors' theories based on whether or not humans are ``super-minds" capable of processing information not Turing-computable.

Nowadays, it is accepted that the nervous system, including the spinal cord and the neocortex, control behavior. In a simplified picture, the nervous system could be viewed as a device which receives input from various senses and produces output in the form of some action. From a computational point of view, the word ``produces" could be harmlessly substituted by ``computes", in the sense that the mind computes in the way a Turing machine does.  One of the most interesting and challenging tasks in science is to assume this as a possibility in order to understand how these computations are performed. Thus, it is not surprising that the field of Neuroscience attracts many researchers not only from the fields of Biology and Computer Science, but also from the area of Physics.

Contributions to the mind's neural computation require a strong effort towards interdisciplinary collaboration. As Teuscher and Sipper\cite{teuscher} pointed out, there are very few problems stemming  from neural computation on which a theoretical computer scientist can  commence work without further interaction or collaboration with neuroscientists.

Since the artificial model was inspired in the natural, in this paper we will explore the bridge between them on the basis of recent work. Since artificial neural networks have been inspired from their biological counterparts, it is natural to ask about the feedback from the artificial to the biological. We attempt build such a bridge by taking the standard model of ANNs and the way in which ARNN admits arbitrary weights, in order to characterize its computational power by using the complexity of the weights or the timing of the pulses.

Most of what happens in our brain does so without our being aware of it, so that many things that seem easy to us involve highly complex neural circuits and operations. Most practical applications of artificial neural networks are based on a computational model involving the propagation of continuous variables from one processing unit to the next.

Inspired by a simplistic vision of how messages are transferred between neurons, computer scientists invented the artificial computational approach to target a wide range of problems in many application areas. Biological neurons are connected by synapses, which are the links that carry messages between them. Using synapses neurons can carry pulses to activate each other with different threshold values. Neurons are the computational part of the network while links connect different neurons and enable messages to flow among them. Each link is a connection with a weight that affects the carried message in a certain way. In fact each link receives a value from an input neuron, multiplies it by a given weight, and then passes it to another neuron. Between this process several sources of computational power can be identified. This means that the source of the computational power of an NN might have at least three possible sources: (a) one carried by the operations in the neuron itself; (b) the message between neurons involving pulses, action potentials (APs) and timing and (c) the complexity of the neural network weights in terms of their capability to encode information which implies (a) and (b). 

As biologically-inspired devices Artificial Neural Networks (ANNs) have been suggested as a computational theory of mental activity. An ANN is mathematically defined as a weighted directed graph in which each vertex represent a neuron and each arrow a connection.

\textit{\textbf{Definition 1.} An Analog Recurrent Neural Network or ARNN\footnote{\textit{The ARNN dynamics is given by a mapping:
\begin{equation}
\mathbb{F}:\mathbb{R^N}X\{0,1\}^M \rightarrow \mathbb{R}^N
\end{equation}
in which each neuron $x_i$ in time (t+1) behaves as described by the following non-linear equation:
\begin{equation}
x_i(t+1) = \sigma \left( \sum_{j=1}^Na_{ij}x_j(t)+\sum_{j=1}^Mb_{ij}u_j(t)+c_i \right) \quad i = 1,\ldots,N
\end{equation}}} is a finite network of nodes (called neurons) and connections, wherein the synaptic weight associated with each connection (determining the coupling strength of the two end-nodes) is a real (analog) value. }

An ARNN is basically an artificial neural network allowing loops (recurrent) and irrational (even non-Turing computable) numbers as weights (analog).

We consider the question about the possible computational power of the human mind in this paper as a problem of computability in general and tractability or practical computability out of its scope then making only emphasis on the computability power defined by the Turing degrees and beyond.

The key arguments which form the bases of our position are: 

\begin{enumerate}
\item ARNNs are a plausible model of the human brain because they define a wide range of different levels of computational power.
\item At least some characteristics of the human mind arise from the human brain.
\item ARNNs are therefore a plausible model of the human mind.
\item It has been argued that the human mind could be depicted or fully simulated by Turing machines, and others, as it has a ``super-Turing'' computational power.
\item Therefore, \emph{prima facie}, it seems reasonable to explore claims on the human mind's supposed Turing or ``super-Turing'' capacities through an analysis of ARNNs in relation to recent work on the brain's own neural network.
\item That an analysis of ARNNs supposed Turing and ``super-Turing'' capabilities (and other more well established findings) gives rise to a taxonomy of computational capabilities.
\item The taxonomy of computational capabilities gives rise to \emph{prima facie} reasonable hypotheses concerning the human mind. The contribution of this paper to our knowledge is to build up this argument and generate hypotheses binding artificial models to the natural.
\end{enumerate}

With reference to argument number 1, the ARNN model even allows to consider a situation in which the weights and the ARNN become equivalent in power to automaton of lower power (including the Turing machine). However, the ARNN (and a wider generalization proposed in this paper, to be fully explored in another, yet to be published) allows us to consider simultaneously, all possible scenarios.  Claim number 2 is widely accepted in different degrees of engagement. It is the case of the mind/body problem and our claim is based on the consideration of the mind as the brain operation. Hence, no distinction is made between brain and mind. However, it is our assumption that the mind does not subtract any computational power and fully inherits the whole computational power of the brain. If mind adds some computational power to the overall system our arguments still apply as lower bound. By contrast, if the mind subtract computational power, our claims should be reformulated in terms of it and our arguments will lay as upper bound.

Our intent is to discuss the mathematical features which a model should possess, if it is to aspire to certain computable power. Our choice of an ARNN model rests on two considerations: (a) it has been proposed as a model with (potentially) non computable capabilities (provided some requirements are met; see later); (b) as a neural network model, neuroscientists might find it easier to relate it to their own empirical work -most of their research is based on (pseudo)analog values- and therefore they can put forward and test concrete hypotheses to confirm or refute the model. Additionally, the ARNN represents a refinement on what had been typically regarded as the mathematical definition of an ANN, which did not make any distinction between the complexity of the weights in terms of the computational power beyond the seminal work of Minsky. Nevertheless, neural networks which run on digital computers operate on a Turing-computable subset of irrational numbers, a strong restriction that determines \emph{a priori} its computational power. Hence, some enthusiasm generated by important experimental and theoretical results cannot be extended easily to applications because there is no straightforward way to make real numbers available even assuming its possible existence. Digital hardware implementation uses a finite number of bits for weight storage and rational restraint values for firings rates, weights and operations and remain limited to a computational power. Even analog implementations, which are often cited for their ability to implement real numbers easily (such as analog quantities), are limited in their precision by issues such as dynamic range, noise, VLSI\footnote{Very Large Scale Integration are systems of transistor-based circuits into integrated circuits on a single chip. For example, the microprocessor of a computer is a VLSI device.} area and power dissipation problems. Thus, in theory and in practice, most implementations use rational numbers, or at most, a subset of irrational numbers, the Turing-computable ones.

The classical approach of computability theory is to consider models operating on finite strings of symbols from a finite alphabet. Such strings may represent various discrete objects such as integers or algebraic expressions, but cannot represent general real or complex numbers, even though most mathematical models are based on real numbers. The Turing machine\cite{Turing1} and its generalization in the form of the Universal Turing machine (UTM) is the accepted model for computation, while under the Church-Turing thesis, it is considered the authoritative model for effective computation. However, some researchers have proposed other models in which real numbers play the main role in effective computations.

Machines with ``super-Turing" capabilities were first introduced by Alan Turing\cite{Turing2}, which investigated mathematical systems in which an oracle was available to compute a characteristic function of a (possibly) non-computable set. The idea of oracles was to set up a scheme for investigating relative computation. Oracles are Turing machines with an additional tape, called the oracle tape, which contains the answer to some non-computable characteristic functions.  An oracle machine is an abstract machine used to study decision problems. It can be visualized as a Turing machine with a black box, called an oracle, which is able to determine certain decision problems in a single step. However, Turing gave no indication on how such an oracle might be implemented. The Turing machine can write an input on its own tape, then tell the oracle to execute it.  In a single step, the oracle computes its function, erases its input, and writes its output to the tape. It is possible to posit the existence of an oracle, which computes a non-computable function, such as the answer to the halting problem or some equivalent. Interestingly, the halting problem still applies to such machines; that is, although they can determine whether particular Turing machines will halt on particular inputs, they cannot determine whether machines with equivalent halting oracles will themselves halt. This fact creates a hierarchy of machines according to their Turing degree, each one with a more powerful halting oracle and an even more difficult halting problem.

With such a method, an infinite hierarchy of computational power can easily be constructed by positing the existence of oracles that perform progressively more complex computations which cannot be performed by machines that incorporate oracles of lower power. Since a conventional Turing machine cannot solve the halting problem, a Turing machine with a Halting Problem Oracle is evidently more powerful than a conventional Turing machine because the oracle can answer the halting question. It is straightforward to define an unsolvable halting problem for the augmented machine with the same method applied to simpler halting problems that lead to the definition of a more capable oracle to solve that problem. This construction can be continued indefinitely, yielding an infinite set of conceptual machines that are progressively more powerful than a Turing machine. This build the hierarchy of Turing degrees:

\textit{\textbf{Definition 2.} The Turing degree of a subset $A$ of natural numbers is the equivalence class consisting of all subsets of $B$ equivalent to $A$ under Turing reducibility. The degree of a set $A$ is denoted by deg($A$). The least element in the partial order is denoted by $0$ and is the degree of all recursive sets (computable sets).}

In other words, two sets of natural numbers have the same Turing degree when the question of whether a natural number belongs to one can be decided by a Turing machine having an oracle that can answer the question of whether a number belongs to the other, and vice versa. So the Turing degree measures precisely the computability or incomputability degree of a subset $X$. 

\textit{\textbf{Definition 3.} A Turing reduction from a $A$ to $B$ is a reduction which easily computes $A$ assuming $B$, i.e. $A$ is computable by a Turing machine with an oracle for $B$.}

Because a language can be always codified by a subset of natural numbers, in terms of computability theory, sets and languages are equivalent.

\textit{\textbf{Definition 4.} Formally, a language $L$ is recognized by a Turing machine with an oracle $A$ if for every string $s$ the machine using $A$ as an oracle halts on input $s$ if $s$$\in$$L$. A language $B$ is Turing-reducible to a language $A$ if a Turing machine equipped with an oracle for $A$ can decide $B$.}

Models of hyper-computation tend to be of two general types:  One uses oracles or oracles in disguise, and the other uses infinite computation in finite time. Whether or not such machines are physically constructible -most experts believe they are not- studying them improves our understanding of the recursion theory.

On the other hand neural networks have been studied as computational devices. In 1956, Kleene showed\cite{Kleene} how to simulate finite automata using McCulloch and Pitts neurons\cite{Cul-Pit} and proved that when the weights of the networks are constrained to be integers, the languages accepted by them are exactly the regular languages. On the other hand, Minsky\cite{minsky} showed that neural networks with boolean neurons can simulate any finite automaton. More recently, Hava Siegelmann\cite{Sieg2} presented a computation model based on analog recurrent neural networks (ARNN). All this work establish a very strong connection between artificial neural networks and automata theory. Both automaton (including the Turing machine) and neural networks are characterized by the languages they accept and hold the same hierarchy.

Siegelmann offered a proof\cite{Sieg2} in which the set of languages accepted by networks with rational weights are exactly the recursively enumerable languages\footnote{Davis\cite{Davis3} rightly pointed out that even if a subset of non rational numbers is used, namely the set of Turing-computable irrationals, the class of languages recognized by neural networks remains the same, as Siegelmann's proof on the power of networks with rational weights readily extends to nets with computable irrational weights (as Turing already did with his machines).}. Siegelmann showed that ARNNs are strictly more powerful than the Turing machine model in that they can perform computations provably uncomputable by a universal Turing machine. Disregarding the fact that it seems unrealistic for most experts that those ARNN could be someday constructible it is not easy to discard if such devices are already present in nature taking advantage on the physical properties of the universe or, at least, the way in which they can perform computations over its  physical environment\footnote{Another interesting question raises by its own right: if there exists a natural device with such capabilities how might we be restricted to take advantage of the same physical properties in order to build an artificial equivalent device? Much of the defense of the work mentioned above have precisely centered on questions such as if we are taking advantage of the resources we have in nature.}. According to G. Kreisel himself it is an open question whether there is any "natural phenomenon" leading to an uncomputable number\cite{calude}.

These neural networks satisfy some classical constraints of computational theory: the input, output, and transitions, are discrete and finite. But the overall system is not really finite since it uses arbitrary real numbers, and it is known that it is powerful enough to encode all non-recursively enumerable languages.  Consequently, a connection between the complexity of the networks in terms of their information complexity and their computational power had been developed, spanning a hierarchy of computation from the Turing model to a ``super-Turing'' mathematical model.

\section{Modeling the Brain}

A clear difference between the brain and the computer is that a digital computer operates by performing sequential instructions from an input program, whereas there is no clear analogy of such a performance in the human brain. However, since any Turing machine working on several tapes is finally equivalent to a single tape Turing machine, the difference may be functional rather than fundamental. The brain as a fully parallel device, could be reduced to a sequential input under some very significant restrictions (for example being discrete or finite). Nowadays, such questions are the subject of scientific and philosophical debate since we have the computational resources to develop live experiments\footnote{There are at least three important projects currently running: A Linux cluster running the MPI NeoCortical Simulator (NCS), capable of simulating networks of thousands of spiking neurons and many millions of synapses, was launched by Phil Goodman at the University of Nevada. Blue Brain, a 4,000 processor IBM BlueGene cluster, was used to simulate a brain in a  project started in May 2005 in the Laboratory of Neural Microcircuitry of the Brain Mind Institute at the EPFL in Lausanne, Switzerland, in collaboration with lab director Henry Markram. It has  as its initial goal, the simulation of groups of neocortical columns  which can be considered the smallest functional units of the neocortex. Also running is  the NCS, to be combined with Michael Hines' NEURON software. The simulation will not consist of a mere artificial neural network, but will involve much more biologically realistic models of neurons. Additionally, CCortex, a project developed by a private company Artificial Development, planned to be a complete 20-billion neuron simulation of the Human Cortex and peripheral systems, on a cluster of 500 computers: the largest neural network created to date. Different versions of the simulation have been running since June 2003. CCortex aims to mimic the structure of the human brain with a layered distribution of neural nets and detailed interconnections, and is planned to closely emulate specialized regions of the Human Cortex, Corpus Callosum, Anterior Commissure, Amygdala and Hippocampus.}. However, Digital computers were not designed to be models of the brain even when they are running neural networks to simulate its  behavior within their own computational restrictions. Most fundamental questions are however related to its computational power, in both senses: time/space complexity and degree of solvability.  Most computational brain models programmed to date are in fact strictly speaking, less powerful than a UTM. Researchers such as Stannett\cite{stannett} have speculated that ``if biological systems really do implement analog or quantum computation, or perhaps some mixture of the two, it is highly likely that they are provably more powerful computationally, than Turing machines. This statement implies that true human intelligence cannot be implemented or supported by Turing machines, an opinion shared by Roger Penrose\cite{penrose3}, who believes mechanical intelligence is impossible since purely physical processes are non-computable. A position strongly criticized by many researchers in the field since its authors first propagated the idea.

However, assuming some kind of relation between the mind and the brain's physical actions, neural networks may be accepted as a model of the human mind operation. Since such a mind/brain relation is widely accepted in different ways and levels, we concern ourselves with the computational power of these devices and the features that such networks must possess.

We will address which from our point of view, goes to the crux of the matter when the question of  the computational power of the brain is  raised, that is, its solvability degree. This means that we do not concern ourselves with  what could be the recipe in which a simulation could run, since if we restrict ourselves to the discussion of artificial neural networks running on actual digital computers, we will be restricted to the lowest computational degree of solvability. From this, it can be easily deduced that if there is a fundamental difference between the architecture of the brain and digital computers, then the efforts of the artificial neural networks to fully simulate the brain either for the purpose of study or reproduction, are destined to have fundamentally different degrees of power.  

Based on certain  references\cite{selmer} as well as our own research, we have identified at least five mathematical descriptions in which ``super-Turing" capabilities have been formally captured: super-tasks\cite{benacerraf,thomson} and accelerated Turing machines, Weyl machines or Zeus machines\cite{boolos}, Trial and Error machines\cite{putnam2}, Non-Standard Quantum Computation, and the Analog Recurrent Neural Networks\cite{Sieg2}. We have also identified other proposals concerning Turing machines with some kind of interaction between them or the environment\cite{wegner,wiedermann,Copeland}, those models provide a basis for the following claims:
\begin{enumerate}
\item Minds are not computers, because (most) thought processes are not.

\subitem (a) It can ``violate" G\"odel's theorem, therefore is not a computing 
machine, a claim made most famously by Lucas\cite{lucas} in 1961.

\subitem (b) Mind can ``solve" the ``Entscheidungsproblem", therefore is not a computing machine. 

\item Minds are computing devices but not of the same power of Turing machines (maybe G\"odel himself\footnote{In his 1951 Gibbs lecture\cite{godel} G\"odel attempts to use incompleteness to reason about human intelligence. G\"odel uses the incompleteness theorem to arrive at the following disjunction: (a) Either mathematics is incompleteable in this sense, that its evident axioms can never be comprised in a finite rule, that is to say, the human mind (even within the realm of pure mathematics) infinitely surpasses the powers of any finite machine, or (b) or else there exist absolutely unsolvable diophantine problems (or absolute undecidable propositions) for which it cannot decide whether solutions exist. G\"odel finds (b) not plausible, and thus he seems have believed that the human mind was not equivalent to a finite machine, i.e., its power exceeded that of any finite machine, the term used originally for Turing machines}).

\subitem (a) There are special operations that occur  in the brain which are not Turing computable, a claim made most famously by Penrose\cite{penrose3}.

\subitem (b) The mind could be a machine but with access to a certain oracle (from an external source or from a previously coded internal source).
\end{enumerate}

All hyper-computational models presented to date are purely theoretical, and we may well ask whether they can actually be implemented in the sense that the universal Turing machine is implemented or pseudo-implemented in a common digital computer. A personal hyper-computer would be no more an implementation of a hyper-computer model, than a personal computer is of a UTM and nobody has physically implemented any kind of hyper-computer. Most models seems to take advantage of (a) a type of oracle device or (b) an infinite number of steps in a finite amount of time.

Jack Copeland has pointed out\cite{copeland2} an interesting fact concerning the way Turing machines work. He stated that Turing machines are closed systems that do not accept input while operating, whereas the brain continually receives input from the environment. Based on this observation, Copeland has proposed the coupled Turing machine which is connected to the environment via one or more input channels\cite{copeland2}. However, as Christof Teuscher and Moshe Sipper have pointed out\cite{teuscher}, any coupled machine with a finite input stream can be simulated by a UTM since the data can be written on the machine's tape before it begins operation. From dynamic systems we often decide almost in an arbitrary way, when a system will be closed in order to handle it. However, the chain of such external systems, potentially infinite (even just by loops) can create a non-linear system which could truly be more complex and maybe more powerful. Some other claims and critics have been made in this regard.

Some other authors claim that the ``super-mentalistic'' perspective is not a scientific one, as it implies the possibility of assigning non-reducible phenomena to some sort of information processing. However, we believe that this fact does not preclude a study on what formal properties can be required from non-Turing models of the human mind. A clarification of these properties would help us understand to what extent ``super-Turing'' models of the mind can or cannot be considered in a scientific way.

\section{A computational model for the human mind}

\subsection{Determining the power of a neural network by the complexity of its weights}

Neural Networks are able to encode information by several ways. If brain's neural network computes  equal or less than the Turing degree of Turing machines without oracles, their weights should code only whole or rational numbers. However, if brain's NN is more powerful it is clear that it is possible to use an oracle to build a neural network that simulates a Turing machine with that oracle in order to recognize any arbitrary language\cite{Sieg2}. This can be done by encoding the oracle into a real number $r$ which the neural network ``consults'' in order to know if a certain input (or a transformation of it) does belong to the language encoded by $r$.

Because we are interested in the set of Turing degrees of the weights of a neural network, and since not always Turing degrees are comparable, it is necessarily to use the notion of a maximal element.

\textit{\textbf{Definition 4.} Let  be a partially ordered set ($A$,$\leq$). Then an element $r$$\in$$A$ is said to be maximal\footnote{Note that the definition for a maximal element is true for any two elements of a partially ordered set that are comparable. However, it may be the case that two elements of a given partial ordering are not comparable as in the case of Turing degrees as Post proved} if, for all $\alpha$$\in$$A$,$r$$\nleq$$\alpha$.}

We are going to take as ``information" to the repository of languages. In Computer Science and Linguistics a language is a set of symbols from an alphabet. As it is well known, every string on an arbitrary alphabet can be encoded into a string on the binary alphabet \{0,1\}. In the same way, a real number with infinite expansion can also be represented by an infinite binary string. Putting together
these facts, a language can be easily encoded into a real number $r$ in the interval
$[0,1)$ taking the $n$ digit of the expansion of $r$ as the value of the
characteristic function of the language applied to the $n$ string in $\{0,1\}^*$
ordered lexicographically. Additionally, this encoding is unique. Then given a language
$L$, $r_L$ is the real number that encodes it.

\textit{\textbf{Definition 5.} A real number $r_L$ is computable if and only if it is the limit of an effectively converging computable sequence of rational numbers. The Turing degree of a real number $r_L$ is defined as the Turing degree of its binary expansion\footnote{it can be seen as a subset of natural numbers (or in fact the whole set of natural numbers encoded in a single real number concatening all). It is evident that not all real numbers are computable (they are also identified as random in Chaitin theory).}.}

\textbf{Example 1.} This definition is quite natural and robust. For instance,  let $P$ be the set of natural numbers defined by  $P$=\{n$\in$$\mathbb{N}$$|$the digit $n$ of the binary expansion of ($\pi$-$3$) is $1$\}. $P$ is evidently computable. Let be  $RN$ a neural net with weights $\pi$ and $\pi$-$n$. Since $\pi$ is Turing-computable and $\pi$-$n$ too then $RN$ will compute only those languages coded by $\pi$ and $\pi$-$n$.

\textbf{Example 2.} However if $h$ is used as a weight, where $h$ is the Chaitin constant $\Omega$ defined as the probability of halting of a Turing machine $M_i$ for an input $x_i$ $RN$ is going to compute definable languages but non-Turing computable. It is evident that $RN$ computes $h$ and those in its same complexity degree. If the set of maximals of $RN$ has only $h$ then the Turing degree of $RN$ is going to be the Turing degree of h, that is $0'$. Then if the human mind is capable to hold $\pi$, $h$ or any other value as a connection between neurons it will determine its computational power by taking the set of maximals of the weight's Turing degrees.

\subsection{Extracting information from the weights}

A more interesting question is how to verify what type of languages a weight has coded in order to determine the computational power of a neural network. That means what language a real number encodes. Since the function of the encoding is bijective it is just needed to analyze one by one the digits of the binary expansion of the real number in order to reconstruct the encoded language. Of course, languages encoded in this way go well beyond recursively
enumerable languages and most of them will go beyond any Turing computable procedure. It is clear that if a weight really encodes a non rational number it is going to be impossible to extract all digits in a finite time and space for most of them. However, if all weights in the brain neural network are finite it will represent an argument for a computationalism position on the human mind (and depending upon the accepted relation between the brain and mind). If a computable irrational number is taken, for example, $\pi$, the language it encodes is clearly recursively enumerable but infinite. However this procedure obscures the rules (of a suitable Turing machine) that generate the language and it is very difficult to see how such rules could be derived from the number itself and then impossible to determine if the substring $n$($\pi$) for the first $n$ digits of $\pi$ is really going to be $\pi$ at the end or any other number sharing the first $n$ digits. Even further, it seems to be impossible to distinguish between a computable or non-computable number by this way. 

Relativization of arbitrary neural networks by using the computational power of their weights by any procedure could allow us to classify them into the well known traditional hierarchies. For this purpose, we will use an oracle to build a neural network that simulates a Turing machine with that oracle to recognize a language. Let $M$ be the neural network we will build. First, we encode the input $w$ into a real number $r$, as we saw in the last section, then we code the oracle given in the same way. Let $o$ be the real number that encodes the oracle. For each entry $i$ coded in $r$, the oracle coded in $o$ is consulted, when the answer is yes, the digit $i$ is $1$, when it is not, the answer is $0$.

Then $M$ has two parts, one a sub-network $N$ will simulate a Turing machine, and the other, named $O$ will simule the oracle machine\cite{Sieg2}. Let $v(i)$ be the oracle answer for the input $i$, we denote by $c$ the concatenation $c=v(0)v(1)...v(i)...v(n)$, and by $d(c)$ a Cantor encoding\cite{Sieg2} of $r$. The Cantor encoding can be avoided since we are not concerned in this paper in time/space complexity. According to Davis\cite{Davis2} it is enough a straight encoding based on length and lexicographical order\footnote{$\epsilon$$\rightarrow$1, a$\rightarrow$2, b$\rightarrow$3, aa$\rightarrow$4, ab$\rightarrow$5, ba$\rightarrow$6, bb$\rightarrow$7, aaa$\rightarrow$8, aab$\rightarrow$9, aba$\rightarrow$10, $\ldots$ For example if $L$ is defined by all the strings that begins with an ``a'' followed by an arbitrary number of ``b''s$L$=\{a, ab, abb, abbb, $\ldots$\}, then the set $S$ will be \{2, 5, 11, 23, $\ldots$\} and the real number encoding $L$ will be $L_r$=0.0100100000100000000000100$\ldots$ (in fact it is a non rational number in this case).} 

Then, the network $M$ can be described\cite{Sieg2} as a composition of two sub-networks: the first, $N$, for which $r_w$ is a weight, receives the input $w$, and after a fixed computation $N$ submit $y_w$ to $O$, where $y$ comes from $w$ after the computation of the Turing machine simulation, the sub-network which performs the oracle machine and has $d(c)$ = $o$ as weight. The output of the sub-network $O$ will depend on a binary activation function\footnote{A sigmoidal type function called the signal function defined as: $signal(x) = 0$ if x$\leq$0 and 1 in other case.}.

The input arrives on two binary input lines. The first is a data line, which is used to carry a binary input signal, when no signal is present, the output is zero. The second is a validation line, which indicates when the data line is active. It takes the value one while the input is present and zero when not. There are also two output lines, that take the same roles. These conventions allow using all external signals to be binary. This gives rise to a taxonomy of computational capacities summarized in the table of the following section.

If any arbitrary neural network $N$ is equivalent to a network $N'$ with two sub-networks $M$ and $O$ that simulate the Turing machine and the oracle component respectively, then the computational power of the whole network $N$ will be determined by the sub-network $O$. Therefore, it makes sense to classify a neural network by means of the Turing degree of its oracle component. The Turing degree of an oracle in a neural network is the set of maximals of the Turing degrees of the weights of the network. The Turing degree of each weight depends on its encoding capacity.

\subsection{Other sources of computational power: spikes and operations}

In recent years, data from neurobiological experiments has made it increasingly clear that biological neural networks, which communicate through pulses, use the timing of these pulses to transmit information and to perform computation. This realization has stimulated a significant growth of research activity in the area of Spiking (or Pulsed) neural networks (SNNs or PNNs), theoretical analysis, as well as the development of new ANN paradigms\cite{judd,maass5}. From the perspective of our main concern in this paper, an important question is: What type of information in terms of complexity, might be carried by an action potential? Each action potential could represent a single bit of information, similar to a serial digital communication channel without error checking. However, the information per action potential may differ according to the function of the neural network. In traditional neural network models the timing of computation steps is usually ``trivialized". Clearly, precisely timing of spikes would allow neurons to communicate much more information than with essentially random spikes. Otherwise, such temporally-coded information is lost as a source of possible computational power in those traditional models. It has been shown that the use of pulse correlations in addition to pulse rates can significantly increase the computational power of a neural network. These timing encodings finally contribute to the overall power of such networks. 

Biologically-inspired computation paradigms take different levels of abstraction when 
modeling neural dynamics. The production of action potentials or spikes has been ab- 
stracted away in many rate-based neurodynamic models, but recently this feature has gained 
renewed interest. In biological neural networks, information is transmitted by the conduction of action potentials along axons, and information processing takes place at the synapses, dendrites, and soma of neurons\cite{koch}. It has been shown that firing correlations play a significant computational role in many biological neural systems. 

Models of biological neural networks in terms of dynamic systems have been studied and formalized by Hodgkin-Huxley\cite{hodgkin} and FitzHugh-Nagumo equations\cite{fitz,nagumo} and others. Recently experimental evidence has accumulated during the last few years, which indicates that many biological neural systems use the timing of single action potentials (or ``spikes") to encode information\cite{abeles,arbib,bialek,rieke,hopfield,gerstner,wagner,sejnowski,singer,softky,thorpe}. Experiments have shown that in vitro biological neurons fire with slightly varying delays in response to repetitions of the same current injection\cite{aertsen}. Even when this behavior may be explained by noisy versions of neural networks, the source for some kind of possible codification in such processes remains. However, noise certainly affects the computational power of networks of spiking neurons for analog input\cite{maassL,orponen}. Nevertheless, empirical experiments in which complex fluctuations exist, as seen for example in EEG\footnote{Electroencephalography is the neurophysiologic measurement of the electrical activity of the brain by recording from electrodes placed on the scalp or, in special cases, on the cortex. The resulting traces are known as an electroencephalogram (EEG) and represent so-called brain-waves.} signals that are generally taken as noise, may be indicative of complex dynamics for processing information similar to that of pulse propagation networks or spiking neurons, maybe continuously over time.

A central theme in this paper is the coding capabilities of neural networks in which its computational power lies. Most accepted descriptions of the human neural network involve finite and discrete quantities, but some of them remain as continuous variables from one processing. Computer scientists have been studying ANNs for many years. Although ANNs were inspired by real biological networks, typical ANNs do not model a number of aspects of biology that may prove to be important. Real neurons, as we have seen for example, communicate by sending out little spikes of voltage called action potentials (APs). ANNs, however, do not model the timing of these individual APs. Instead, ANNs typically assume that APs are repetitive, and they model only the rate of that repetition. However some of the computational power of a biological neural network could be derived from the precise timing of the individual APs or other properties inherent to the biological neural network that it is not fully right-modeled. Regular ANNs could never model such a possibility defined as their currently are.

Additionally, all ANNs running over digital machines are incapable of simulating any possible analogical signal or full irrational encodings if they were present in human brain's architecture. Methods for estimating the computational power of neural circuits and their relation to artificial neural networks models have been established. Maass\cite{maass1,maass2,maass3,maass4} and Markram\cite{markram} have recently argued that ``in contrast to Turing machines, generic computations by neural circuits are not digital, and are not carried out on static inputs, but rather on functions of time". These kinds of experiments could ultimately provide   a definitive answer to the critical issues which are the concern of this paper.

These models are also interesting because even if a neural network is constrained to simple weights, namely whole or rational numbers, it is possible to achieve an extra power from the timing process, encoding what is not possible to encode in the weights. Since they are similar in power to analog signal processing, they can be compared to the traditional hierarchies we have already explored. Even when a mathematically rigorous analysis of the computational power of networks of spiking neurons has so far been missing, they are equivalent to the levels of such hierarchies, simply by replacing weights encodings with pulse timing encodings. In other words, the computational power is transferred from weights to spikes, and in the presence of a mixed model with both weights and spikes encodings, its final computational power will be the power of the most powerful. Both pulse frequencies and correlations which are computationally relevant can be seen as operations involving potentially any possible real numbers in principle equivalent to a weight encoding. Hence, the overall power of these PNNs or SNNs is determined by the maximal Turing degree of the union of the Turing degrees of both weights and time encodings. It has also been shown that SNNs are, with regard to the number of neurons that are needed, computationally more powerful than other neural network models. These models are obviously Turing reducible when variables (the input and all internal operations) are restricted to Turing-computable values if such model is close under its operations. However when arbitrary values are allowed then they can compute possible non-Turing functions. Maass\cite{maass97} has shown that, with regard to temporal coding of real-valued variables any continuous function can be approximated arbitrarily closely. In this case, the model depends on the continuity of time. 

On the other hand, since it is possible to compute $\sqrt2$ from simple operations between whole numbers (as the hypotenuse of a Pythagorean triangle with unitary sides) and even to make proofs of irrationality without calculating their decimal expansion, it might be possible to prove if brain's neural network can hold non-computable numbers if it is able to perform some special operators where the discrete operations, such as primitive recursion or bounded sums and products, are replaced by operations on continuous functions, such as integration or linear integration\cite{mu}. The sets of those real functions\cite{silva} can be definable by a general-purpose analog computer\cite{shannon} or GPAC, which is a general computing model evolving in continuos time\footnote{\cite{costa1,costa3} The class of $R$-recursive functions is very large. It contains many traditionally non-computable functions, such as the characteristic functions of sets of the arithmetical hierarchy\cite{moo,odi}. Experimental proofs beyond this level could be more difficult if not impossible since the construction of a sequence of real numbers which can not be computably diagonalized is used to prove that there are continuous functions without a Turing degree.}. Then even if weights and spikes are restricted to Turing-computable functions, the brain neural network might be not closed if at least an operation is not recursive possibly (see \cite{costa2} and \cite{costa3}) taking the involved values  to non-Turing degree at some point of the computation. Traditional functional operations like Sequential Composition, Parallel Composition or definition by Cases, Primitive Recursion or simply Recursion, Bounded Recursion, Partial Projection, Cut-off Subtraction and the order functions like the Minimization or Projection and Bounded minimization take computable functions into computable functions. By contrast, It is well known that limits take computable functions beyond the class defined by the power of Turing machines. For a more detailed description of such operations see \cite{burgin}. The power of human brain operations remains unknown, but even if non-traditional operations were possible both theoretical and empirical evidence seem to be hard or impossible since the verification of non-Turing operators require non-Turing inputs and outputs which seems to be undistinguishable from those Turing-computable. Nevertheless neurological tests could be designed in order to achieve some advance in this direction.

One could suppose that given the evolution of neural networks models, several better approaches to the biological neural network model could be expected. Other possible sources could be unrepresented in current artificial models. Experimental and theoretical research in the field should continue to relate it more to the biological model. Maass\cite{maass97} suggests SNNs as the third generation of the artificial neural network model. Each one seems to offer better approaches to the experimental evidence of the brain's own neural network.

Now, we can build a simplified hierarchy of computational power regarding the two possible sources explored in this paper (weights and spikes, which are equivalent in terms that they can be replaced one for the other preserving the whole complexity of the NN):

\begin{center}
\begin{tabular}{|p{1.6in}|p{2.5in}|} \hline
\em Neural Network Architecture &\em Computational Power\\\hline
$\mathbb{R}$ Non-computable numbers & Turing machine with an oracle depending on the maximal Turing degree of of the weights: $w_1$,$w_2$, $\ldots$, $n$ $\in$ $\mathbb{R}$ and pulses $p_1$,$p_2$, $\ldots$, $m$ $\in$ $\mathbb{R}$, not always in the Arithmetical Hierarchy \\\hline
$\mathbb{Q}$ Rational numbers & at most Turing machine \\\hline
$\mathbb{Z}$ Whole numbers & at most bounded automata \\\hline
\end{tabular}
\end{center}

At the bottom level, we have neural networks computing Turing-computable languages if the oracle is empty. On the other hand, weight's and spikes' complexity could be determined -assuming that the neural network is closed under isomorphism- then they can be decoded into an oracle which will determine its computational power. If the neural network perform not closed operations under isomorphism, the neural network will compute in a range determined by the bounds of such operations, possibly determining an interval within this simplified hierarchy.

\section{Discussion and Conclusion}

Even when the answer concerning the computational power of the human mind should come from neurophysiological and interdisciplinary research, including the correspondence between the natural and the artificial neural network models, the question remains of mathematical and philosophical interest as it is possible to explore all possible scenarios inside the computability theory. We add that if we were able to determine some key properties of the physical universe and the mind we could determine the computational power of the mind, among them:

\begin{itemize}

 \item if our physical universe is discrete or continuous, and in which case

\item if it can hold non-computable numbers;

\item if it can hold non-computable operations;

\item if the brain inherits those properties;

\item if the mind can take advantage of them, which depends on the precise relationship between the mind and the physical operation of the brain.

\end{itemize}

As it can be seen, these are not small requirements. However once we can ascertain some fundamental properties of the mind it could be possible to know to which computational level it belongs by its maximal Turing degree. The value of this paper lies in its effort to restore some of the claims which have been made and proven for artificial neural networks to the natural model and the problem of determining the mind's computational power. Then, it is not just about the way in which a neural network can achieve one or more computational powers, but also which features a brain should have to potentially take its power to one level or another: either less, equal or more powerful than Turing machines.

Then, if every language with a given alphabet can be encoded and potentially extracted into a real number (from weights or spikes), this real number can be used as the main component of a neural network that simulates an oracle Turing machine whose oracle is the characteristic function of the language. Therefore  by determining the neural network encoding capability and knowing the type of operations permitted in it, its computational power could be determined. 

We would like to conclude, by remarking that a full model of the mind has a very difficult task ahead. We hope that this paper has succeeded in shedding light on the current research and ways to build certain bridges from the artificial to the natural model, as a possible way to eventually determine the computational power of the human mind taking in consideration all possible sources of power.

\section*{Acknowledgments}
The authors would like to thank the editors and referees for very helpful comments during the preparation of this paper.

\bibliographystyle{latex8}

\begin{thebibliography}{1}

\bibitem{abeles} M. Abeles, Corticonics. \emph{Neural circuits in the cerebral cortex}, Cambridge, England: Cambridge University Press, 1991.

\bibitem{aertsen} A. Aertsen, M. Erb, and G. Palm, \emph{Dynamics of functional coupling in the cerebral cortex: an attempt at a model-based interpretation}, Physica D, 75, 103-"128, 1994.

\bibitem{arbib} M. Arbib (Ed.) \emph{The Handbook of Brain Theory and Neural Networks}, 2nd Edition. Cambridge and MIT Press. 1072-1076, 2002.

\bibitem{benacerraf} P. Benacerraf, \emph{Tasks, super-tasks, and the modern elastics}, J. Philos. 59 765-784, 1962.

\bibitem{selmer} S. Bringsjord and M. Zenzen, \emph{Superminds, People harness hypercomputation and more}, Kluwer, 2003.

\bibitem{bialek} W. Bialek, , F. Rieke, R. Steveninck, de Ruyter van, and Warland, D. \emph{Reading a neural code}. Science, 252, 1854-1857.

\bibitem{boolos} G. Boolos, J. P. Burgess and R. C. Jeffrey, \emph{Computability and Logic}, Cambridge University Press, 2002.

\bibitem{burgin} Mark Burgin, \emph{Super-Recursive Algorithms (Monographs in Computer Science)}, Springer, 2004.

\bibitem{calude} Cristian S. Calude,  \emph{Information and Randomness : An Algorithmic Perspective (Texts in Theoretical Computer Science. An EATCS Series)}, Springer, 2002.

\bibitem{costa2} Manuel Campagnolo, Cristopher Moore, and Jose Felix Costa. \emph{Iteration, inequalities, and differentiability in analog computers}, Journal of Complexity, 16(4):642--660, 2000.

\bibitem{Copeland} B. B. Jack Copeland, \emph{Hypercomputation}, Minds and Machines 12: 461-502, Kluwer Academic Publishers, 2002. 

\bibitem{copeland1} B. B. Jack Copeland, and D. Proudfoot. \emph{Alan Turing's forgotten ideas in computer science}. Scientific American 280, 4, 77-81, 1999.

\bibitem{copeland2} B. B. Jack Copeland and R. Sylvan, R. \emph{Beyond the universal Turing machine}. Australasian J. Philosophy 77, 1, 46-66, 1999. 

\bibitem{Costa} J. F\'elix Costa, \emph{On Hava (Eve) T. Siegelmann's book Neural Networks and Analog Computation\/}'', Universidad T\'ecnica de Lisboa-Birkhauser, 1999. 68--83 University of Nevada, Reno.

\bibitem{costa2} J. Mycka and F\'elix Costa, Jos\'e. emph{Real recursive functions and their hierarchy}, Journal of Complexity, 20(6):835-857, 2004.
 
\bibitem{costa3} J. Mycka and F\'elix Costa, Jos\'e. \emph{What lies beyond the mountains, computational systems beyond the Turing limit}, European Association for Theoretical Computer Science Bulletin, 85:181-189, 2005.

\bibitem{Davis1} M. Davis, \emph{Computability and Unsolvability}, Dover Publications, 1958.

\bibitem{Davis2} M. Davis, \emph{The Undecidable}, Dover Publications, 2004.

\bibitem{Davis3} M. Davis, \emph{The Myth of Hypercomputation, Alan Turing: Life and Legacy of a Great Thinker}, Springer, 2004.

\bibitem{fitz} R. FitzHugh, \emph{Impulses and physiological states in models of nerve membrane}, Biophysical Journal, 1:445–466, 1961. 

\bibitem{gerstner} W. Gerstner, R. Kempter, J. van Hemmen and H. Wagner, \emph{A neuronal 
learning rule for sub-millisecond temporal coding}, Nature, 383, 76-78, 1996. 

\bibitem{godel} K. G\"odel,  \emph{Some Basic Theorems on the Foundations of Mathematics and Their Implications}, in Feferman, et al., 1995, 304-323, 1951.

\bibitem{hodgkin} A.L. Hodgkin, A.F. Huxley, \emph{A quantitative description of membrane current and its application to conduction and excitation in nerve}, J. Physiol. 117 500-544, 1952.

\bibitem{hopfield} J.J. Hopfield, \emph{Pattern recognition computation using action potential 
timing for stimulus representation}, Nature, 376, 33-36, 1995.

\bibitem{hopfield1} J.J. Hopfield, \emph{Neural networks and physical systems with emergent collective computational abilities}, Proc. Natl. Acad. Sci. USA 79 2554–2558, 1982.

\bibitem{judd} K. Judd, K. Aihara, \emph{Pulse Propagation Networks: A Neural Network Model That Uses Temporal Coding by Action Potentials}, Neural Networks, Vol. 6, pp. 203-215, 1993

\bibitem{Kleene} S.C. Kleene, \emph{Representation of events in nerve nets and finite automata}. In C.E. Shannon and J. McCarthy, editors, Automata Studies, pages 3-42. Princeton University Press, Princeton, NJ, 1956.

\bibitem{koch} C. Koch, \emph{Biophysics of Computation: Information Processing in Single Neurons€}, Oxford University Press, 1999.

\bibitem{lucas} JR. Lucas, \emph{Minds, Machines and Gödel}. Philosophy, 36:112-127, 1961.

\bibitem{maass1} W. Maass, R. A. Legenstein, and N. Bertschinger. \emph{Methods for estimating the computational power and generalization capability of neural microcircuits}. In L. K. Saul, Y. Weiss, and L. Bottou, editors, Advances in Neural Information Processing Systems, volume 17, pages 865-872. MIT Press, 2005.

\bibitem{maassL} W. Maass, \emph{Lower bounds for the computational power of spiking neurons. Neural Computation}, 8:1–40, 1996. 

\bibitem{maass2} W. Maass. \emph{On the computational power of neural microcircuit models: Pointers to the literature}. In José R. Dorronsoro, editor, Proc. of the International Conference on Artificial Neural Networks -- ICANN 2002, volume 2415 of Lecture Notes in Computer Science, pages 254-256. Springer, 2002.

\bibitem{maass3} W. Maass. \emph{Neural computation: a research topic for theoretical computer science? Some thoughts and pointers}. In Rozenberg G., Salomaa A., and Paun G., editors, Current Trends in Theoretical Computer Science, Entering the 21th Century, pages 680-690. World Scientific Publishing, 2001.

\bibitem{maass4} W. Maass. \emph{Neural computation: a research topic for theoretical computer science? Some thoughts and pointers}. In Bulletin of the European Association for Theoretical Computer Science (EATCS), volume 72, pages 149-158, 2000.

\bibitem{maass5} W. Maass and C. Bishop (Eds), \emph{"Pulsed Neural Networks"}, MIT Press 1999. 

\bibitem{maass97} W. Maass \emph{Networks of Spiking Neurons: The Third Generation of Neural Network Models}, Neural Networks, Vol. 10, No. 9, pp. 1659-1671, 1997.

\bibitem{maass05} T. Natschia, W. Maass, \emph{Dynamics of information and emergent computation in generic neural microcircuit models}, Neural Networks 18, 1301–1308, 2005.

\bibitem{markram} W. Maass, T. Natschlager, and H. Markram, \emph{Real-time computing without stable states: A new framework for neural computation based on perturbations}, In Neural Computation, 
volume 14, pages 2531–2560, 2002. 

\bibitem{orponen} W. Maass, and P. Orponen, \emph{On the effect of analog noise on discrete-time analog computations}, Advances in Neural Information Processing 11 Systems 9, 1997, 218–224; journal version: Neural Computation 10, 1998. detailed version see http://www.math.jyu.fi/?orponen/papers/noisyac.ps .

\bibitem{nagumo} J. Nagumo, S. Arimoto, and S. Yoshizawa, \emph{An active pulse transmission line simulating nerve axon}. Proceedings of the IRE, 50:2061–2070, 1962.

\bibitem{Cul-Pit} W.S. McCulloch and W. Pitts. \emph{A logical calculus of the ideas immanent in nervous activity'}, Bulletin of Mathematical Biophysics, 5:115-133, 1943.

\bibitem{moo} C. Moore, \emph{Recursion theory on real-time language recognition by analog computers}. Theoretical Computer Science, 162:23-44, 1996.

\bibitem{mu} Jerzy Mycka. \emph{mu-Recursion and infinite limits}, Theoretical Computer Science, 302:123--133, 2003.

\bibitem{minsky} M. Minsky, \emph{Computation: Finite and Infinite Machines}, Prentice Hall, New york, London, Toronto, 1967.

\bibitem{odi} P. Odifreddi, \emph{Classical Recursion theory}. Elsevier, 1989.

\bibitem{penrose1} R. Penrose, \emph{The Emperor's New Mind Concerning Computers, Minds, and the Laws of Physics}, Oxford: Oxford University Press, 1989.

\bibitem{penrose3} R. Penrose, \emph{Shadows of the Mind: A Search for the Missing Science of Consciousness}, Oxford: Oxford University Press, 1994.

\bibitem{Post} E. Post, ``Recursively enumerable sets of positive integers and their
decision problems'', \emph{bulletin of the American Mathematical Society}, 50:284--316.

\bibitem{putnam1} H. Putnam, \emph{Renewing Philosophy}, Cambridge, MA: Harvard University Press, 1992.

\bibitem{putnam2} H. Putnam, \emph{Trial and Error Predicates and the Solution of a Problem of Mostowski}, Journal of Symbolic Logic 30, pp. 49–57, 1965.

\bibitem{rieke} F. Rieke, D. Warland, R.R. von Steveninck, and W. Bialek, \emph{Spikes: Exploring the Neural Code}, MIT Press, Cambridge, USA, 1997. 

\bibitem{searle} J.R. Searle, \emph{Minds, Brains, and Programs}, Behavioral and Brain Sciences,  3, 417-457, 1980.  

\bibitem{sejnowski} T.J. Sejnowski. \emph{Neural pulse coding}. In W. Maass and C.M. Bishop, editors, Pulsed Neural Networks, The MIT Press, 1999. 

\bibitem{silva} Daniel Silva Graa. \emph{Some recent developments on Shannon's General Purpose Analog Computer}, Mathematical Logic Quarterly, 50(4-5):473--485, 2004.

\bibitem{singer} W. Singer, \emph{Neuronal synchrony: A versatile code for the definition of 
relations}, Neuron, 24, 49-65, 1999.

\bibitem{softky} W. Softky, \emph{Simple codes versus efficient codes}, Curr. Opin. Neurobiol., 
5, 239-247, 1995.

\bibitem{shannon} C. Shannon, \emph{Mathematical theory of the differential analyser}. J. Math. Phys. MIT, 20:337-354, 1941.

\bibitem{Sieg1} H.T. Siegelmann, \emph{Computation Beyond the Turing Limit}, Science, 268:545-548, April 28 1995.

\bibitem{Sieg2} H.T. Siegelmann, \emph{Neural Networks and Analog Computation, beyond the Turing Limit}, Birkhauser, 1999.

\bibitem{Sieg-Son2} H.T. Siegelmann and E.D. Sontag, \emph{Analog computation via neural networks}, Theoretical Computer Science, 131:331-360, 1994.

\bibitem{Sieg-Son3} H.T. Siegelmann and E.D. Sontag, \emph{On computational power of neural networks}, Journal of Computer and Systems Sciences, 50(1):132.150, 1995.

\bibitem{costa1} Daniel Silva Graa and Jose Felix Costa, \emph{Analog computers and recursive functions over the reals}, Journal of Complexity, 19(5):644--664, 2003.

\bibitem{stannett} M. Stannett. \emph{X-machines and the halting problem: Building a super Turing machine}, Formal Aspects of Computing 2, 4, 331-341, 1990. 

\bibitem{rall} W. Rall, \emph{Theoretical significance of dendritic tree for input-output relation}, in: R.F. Reiss (Ed.), Neural Theory and Modeling, Standford University Press, Stanford, pp. 73-97, 1964.

\bibitem{teuscher} Christof Teuscher and Moshe Sipper, \emph{Communications of the ACM}, Vol. 45, No. 8, 2002.

\bibitem{thorpe} S. Thorpe, F. Fize, and C. Marlot, \emph{Speed of processing in the human visual system,} Nature, 381, 520-522, 1996.

\bibitem{Turing1} A.M. Turing, \emph{On computable numbers, with an application to the Entscheidungsproblem}, Proc. London Math. Soc. 42 (2) (1936) 230-265. A correction in 43 (1937) 544-546 (reprinted in: M. Davis, (Ed.). The Undecidable: Basic Papers on Undecidable Propositions. Unsolvable Problems and Computable Functions, Raven, New York, 1965, pp. 115-154) Page numbers refer to the 1965 edition.

\bibitem{thomson} J. Thomson, 1954-55, \emph{Tasks and Super-Tasks}, Analysis, XV, pp. 1-13; reprinted in Salmon, 1970.

\bibitem{Turing2} A.M. Turing, \emph{Systems of logic based on ordinals}, Proc. London Math. Soc. 45 (2) (1939) 161-228 (reprinted in: M. Davis (Ed.). The Undecidable: Basic Papers on Undecidable Propositions. Unsolvable Problems and Computable Functions, Raven, New York, pp. 154-222, 1995.

\bibitem{wagner} K. Wagner and T.M. Slagle. \emph{Optical competitive learning with VLSI liquid-crystal winner-take-all modulators}. Applied Optics, 32(8):1408–1435, 1993.

\bibitem{wegner} P. Wegner and D. Goldin,  \emph{Interaction, Computability, and Church's Thesis}, University of Massachusetts at Boston, 1999. 

\bibitem{wiedermann} J. Wiedermann and J. van Leeuwen,  \emph{Emergence of a Super-Turing Computational Potential in Artificial Living Systems}, J. Kelemen and P. Sosik (Eds.): ECAL 2001, LNAI 2159, pp. 55–65, Springer-Verlag Berlin Heidelberg, 2001.

\end{thebibliography}

\end{document}